\pdfoutput=1

\documentclass[11pt]{article}

\usepackage[]{EMNLP2022}
\usepackage{times}
\usepackage{algorithm}
\usepackage{algorithmic}
\usepackage{latexsym}
\usepackage{amssymb}
\usepackage{tabularx}
\usepackage{float}
\usepackage{amsmath}
\usepackage{graphicx}
\usepackage{bbding}
\usepackage[T1]{fontenc}

\usepackage[utf8]{inputenc}

\usepackage{microtype}

\usepackage{inconsolata}

\title{Supervised Prototypical Contrastive Learning for Emotion Recognition in Conversation}

\author{Xiaohui Song$^{1,2\star}$, Longtao Huang$^{3}$, Hui Xue$^{3}$ \and Songlin Hu$^{1,2\dag}$ \\
$^{1}$Institute of Information Engineering, Chinese Academy of Sciences\\ 
$^{2}$School of Cyber Security, University of Chinese Academy of Sciences \\
$^{3}$Alibaba Group \\
  \texttt{\{songxiaohui,husonglin\}@iie.ac.cn}, \texttt{\{kaiyang.hlt, hui.xueh\}@alibaba-inc.com}
}

\newcolumntype{C}[1]{>{\centering\arraybackslash}p{#1}}
\begin{document}
\maketitle
\begin{abstract}
Capturing emotions within a conversation plays an essential role in modern dialogue systems. However, the weak correlation between emotions and semantics brings many challenges to emotion recognition in conversation (ERC). Even semantically similar utterances, the emotion may vary drastically depending on contexts or speakers. In this paper, we propose a \textbf{S}upervised \textbf{P}rototypical \textbf{C}ontrastive \textbf{L}earning (\textbf{SPCL}) loss for the ERC task. Leveraging the Prototypical Network, the SPCL targets at solving the imbalanced classification problem through contrastive learning and does not require a large batch size. Meanwhile, we design a difficulty measure function based on the distance between classes and introduce curriculum learning to alleviate the impact of extreme samples. We achieve state-of-the-art results on three widely used benchmarks. Further, we conduct analytical experiments to demonstrate the effectiveness of our proposed SPCL and curriculum learning strategy. We release the code at \href{https://github.com/caskcsg/SPCL}{https://github.com/caskcsg/SPCL}.

\let\thefootnote\relax\footnotetext{$^{\star}$Work done during internship at Alibaba Group.}
\let\thefootnote\relax\footnotetext{$^{\dag}$ Corresponding Author.}

\end{abstract}
 
\section{Introduction}

\begin{figure}
    \centering
    \includegraphics[width=0.5\textwidth]{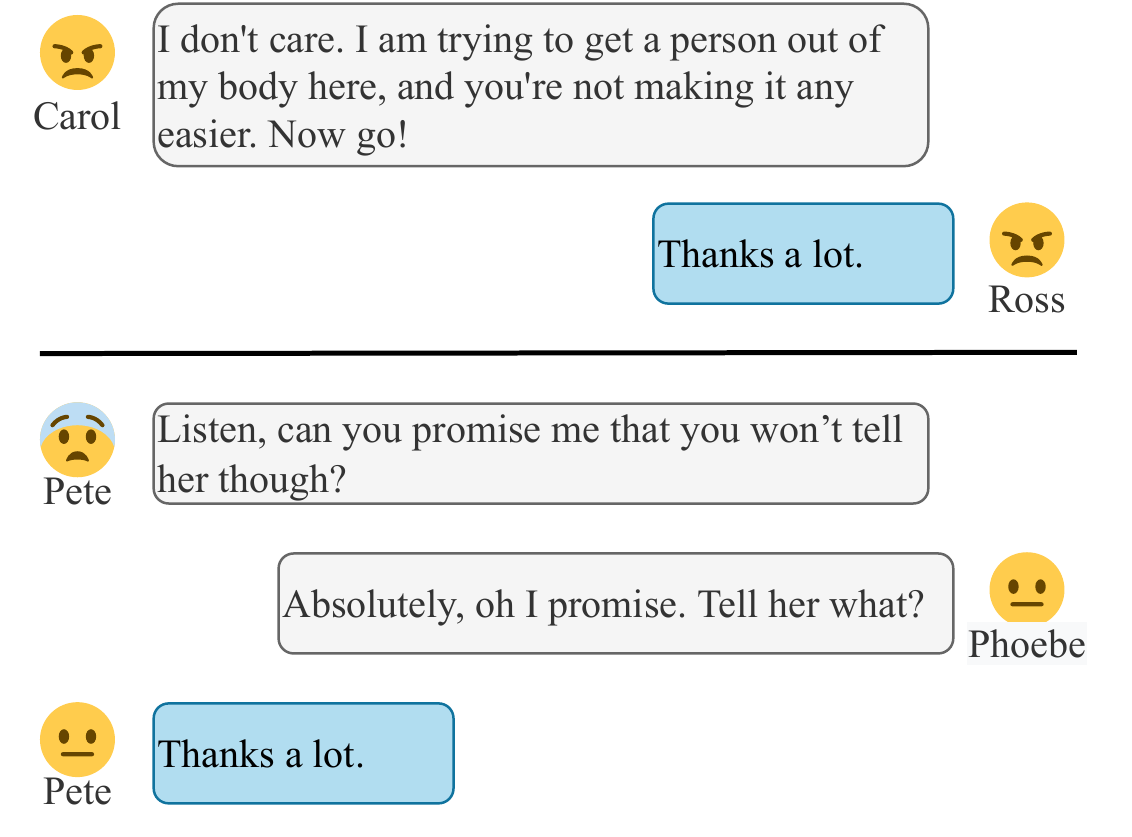}
    \caption{Examples of emotion recognition in conversation. The same utterance "Thanks a lot" can convey different emotions in different contexts.}
    \label{fig:example}
\end{figure}

With the development of online social networks, capturing emotions during conversations has gained increasing attention in both academia and industry\citep{li2020multi,shen2021directed,wang2020contextualized,ghosal2020cosmic, song2022emotionflow,zhu2021topic}. Emotion recognition in conversation (ERC) is critical in many scenarios, such as chatbots, healthcare applications, mining opinions on social media, and so on\citep{poria2019emotion}. The ERC task aims to identify different emotions at each turn within a conversation based on the transcript. A conversation often contains several speakers and runs several turns; thus, emotions can vary drastically during the conversation. Compared to traditional text classification tasks, figuring out emotions needs not only one turn of textual utterance but also contextual information. An example of ERC is illustrated in Figure \ref{fig:example}.

\begin{figure}[h]
    \centering
    \includegraphics[width=0.48\textwidth]{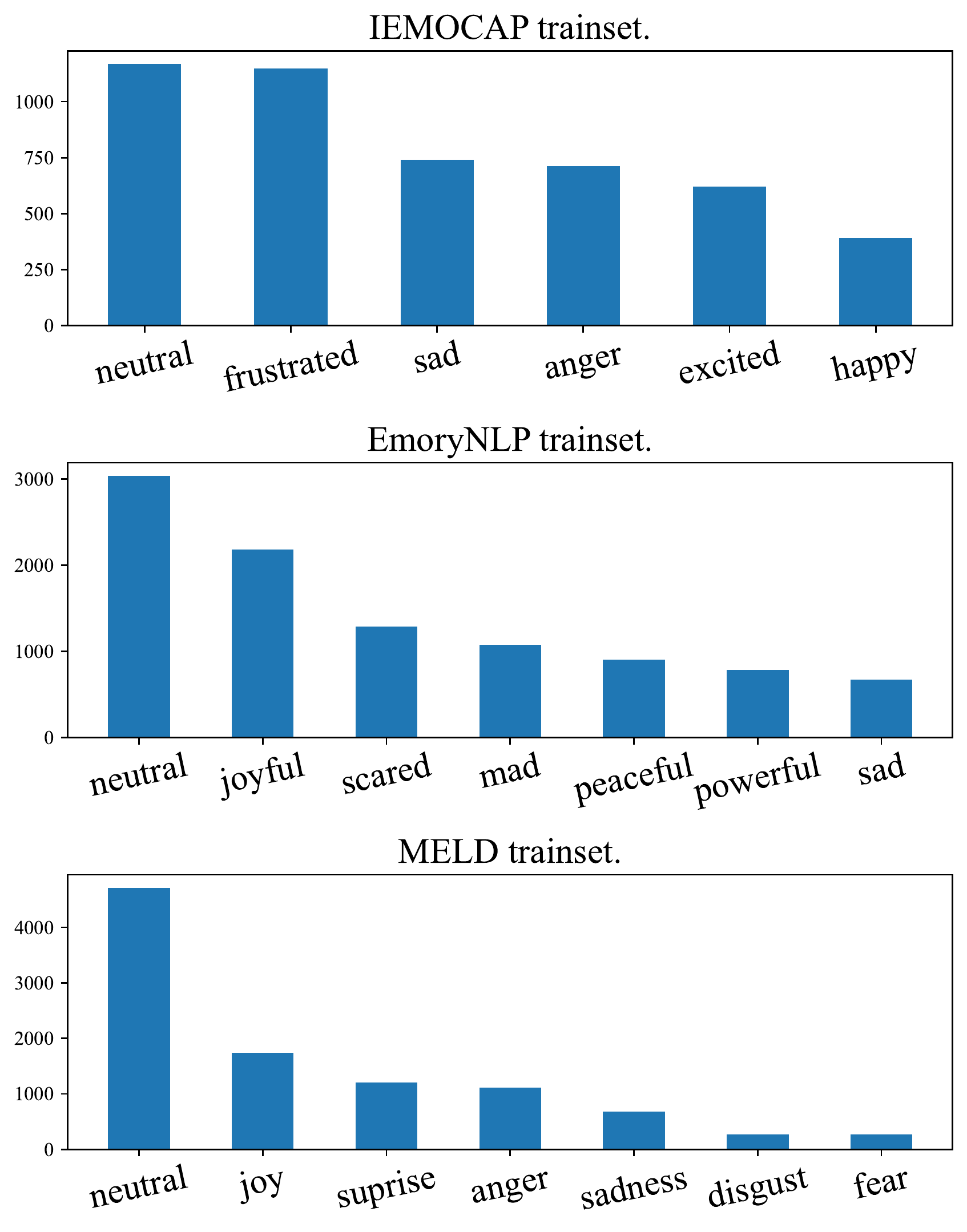}
    \caption{Emotion distributions of the three datasets.}
    \label{fig:my_label}
\end{figure}

Contrastive learning applied to self-supervised representation learning has seen a resurgence in recent years.  \citet{khosla2020supervised} extended the self-supervised batch contrastive approach to the fully-supervised setting and show outperformance over cross-entropy loss in several benchmarks. Although CoG-BART\cite{li2021contrast} has demonstrated the effectiveness of supervised contrastive learning (SCL) in the ERC task, there are still two issues worth to solve when building an ERC model with SCL: (1) As illustrated in Figure \ref{fig:my_label}, existing ERC datasets are often class-imbalanced and samples may not be able to meet appropriate positive/negative samples within a mini-batch. (2) Existing ERC datasets are usually collected in a multi-modal manner. There are some conversations whose textual information is insufficient to distinguish emotions. Training a textual ERC model with those extreme samples may lead to performance degradation.

For the first issue, we propose a \textbf{S}upervised \textbf{P}rototypical \textbf{C}ontrastive \textbf{L}earning (\textbf{SPCL}) loss, which integrates Prototypical Network\cite{snell2017prototypical} and supervised contrastive learning. SPCL maintains a representation queue for each category. At each training step, SPCL samples a certain number of representations from these queues as the support set and calculates a temporary prototype vector for each emotion category. These prototype vectors are used as samples of the corresponding class to compute the loss. SPCL ensures that each sample has at least one positive sample of the same category and negative samples of all other categories within a mini-batch. Experiments show that SPCL can work well in class-imbalanced scenarios and is less sensitive to the training batch size.

To alleviate the performance degradation caused by extreme samples, we combine curriculum learning\cite{bengio2009curriculum} with contrastive learning. We design a distance-based difficulty measure function. By sorting the training data via this function, we can schedule the training data in an \textit{easy-to-hard} fashion. Experimental results demonstrate the effectiveness of our proposed curriculum learning strategy. Finally, we utilize SimCSE\cite{gao2021simcse}, a pretrained language model trained with contrastive learning as our backbone model. Combining our proposed SPCL loss and curriculum learning strategy, we reach state-of-the-art results on three widely used benchmarks. In summary, our contributions can be concluded as follows:
\begin{itemize}
    \item We propose a \textbf{S}upervised \textbf{P}rototypical \textbf{C}ontrastive \textbf{L}earning (\textbf{SPCL}) loss for the ERC task, which can perform supervised contrastive learning efficiently on class-imbalanced data and has no need for large batch size.
    \item To the best of our knowledge, we are the first to combine supervised contrastive learning and curriculum learning for the ERC task.
    \item We achieve state-of-the-art results on three widely used benchmarks. Experimental results further demonstrate the effectiveness of our proposed SPCL loss and curriculum learning strategy.
\end{itemize}

\section{Related Work}

\subsection{Emotion Recognition in Conversation}
Most existing approaches focus on context modeling. They can be divided into sequence-based, graph-based, and knowledge-enhanced methods. Sequence-based methods consider contextual information as utterance sequences. ICON\cite{hazarika2018icon} and HiGRU\cite{jiao2019higru} both use the gated recurrent unit to capture the context information. DialogRNN\cite{majumder2019dialoguernn} is a recurrence-based method that models dialog dynamics with several RNNs. DialogueCRN\cite{hu2021dialoguecrn} introduces multi-turn reasoning modules to model the ERC task from a cognitive perspective. CoMPM\cite{lee2021compm} models the context and speaker's memory via pretrained language models. For those graph-based methods, DialogGCN\cite{hu2021dialoguecrn} and RGAT\cite{ishiwatari2020relation} build a graph upon the utterances nodes. ConGCN\cite{zhang2019modeling} trades both speakers and utterances as nodes and builds a single graph upon the whole ERC dataset. DAG-ERC\cite{shen2021directed} uses a directed acyclic graph (DAG) to model the intrinsic structure within a conversation. Knowledge-enhanced methods\cite{zhong2019knowledge,zhu2021topic,ghosal2020cosmic,zhang2020knowledge} usually utilize external knowledge from ATIMOC\cite{sap2019atomic} or ConceptNet\cite{liu2004conceptnet}. Besides individual models, several frameworks have also been proposed. \citet{yang2021hybrid} developed an ERC-oriented hybrid curriculum learning framework and \citet{ijcai2022p562} proposed a speaker-guided encoder-decoder framework, formulating the modeling of speaker interactions as a flexible component.

\subsection{Contrastive Learning}

In the field of natural language processing, SimCSE\cite{gao2021simcse} is a state-of-the-art contrastive learning framework for generating sentence embeddings, it can learn from unlabeled sentences or annotated pairs from natural language inference datasets. \citet{khosla2020supervised} extend the self-supervised batch contrastive approach to the fully-supervised setting
to make full use of label information. \citet{yeh2021decoupled} let the contrastive learning get rid of the dependence on large batch size. CoG-BART\cite{li2021contrast} adapts supervised contrastive learning to make different emotions mutually exclusive to identify similar emotions better.
\section{Methodology}

\subsection{Definition}

Given a collection of all speakers $\mathcal{S}$, an emotion label set $\mathcal{E}$ and a conversation $\mathcal{C}$, our goal is to identify speaker's emotion label at each conversation turn. A conversation is denoted as $[(s_1, u_1), (s_2, u_2), \cdots, (s_N, u_N)]$, where $s_i \in \mathcal{S}$ is the speaker and $u_i$ is the utterance of $i$-th turn. In this paper, we focus on the real-time settings of ERC, in which model can only take previous turns $[(s_1, u_1), (s_2, u_2), \cdots, (s_t, u_t)]$ as input to predict the emotion label $y_t$ of $t$-th turn.

\subsection{Context Modeling}
\label{sec32}
\begin{figure}
    \centering
    \includegraphics[width=0.48\textwidth]{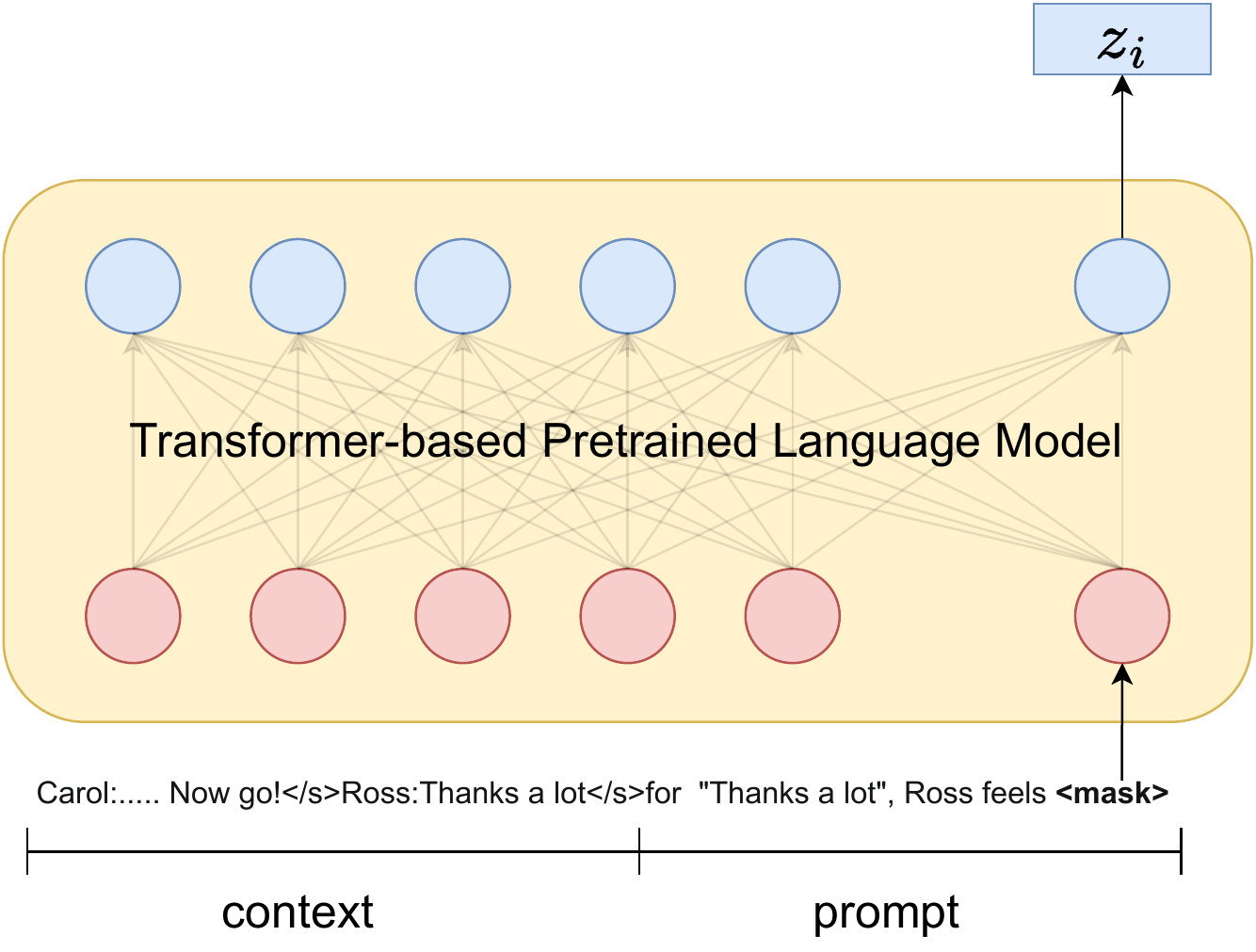}
    \caption{The architecture of our prompt-based context encoder.}
    \label{fig:encoder}
\end{figure}

We build a prompt-based context encoder upon SimCSE\cite{gao2021simcse} to get speaker and context-aware emotion representations. The architecture of the context encoder is illustrated in Figure \ref{fig:encoder}. To calculate representation for $u_t$, we use the most recent $k$ turns of utterances and speakers as context.
\begin{equation}
C_t = [s_{t-k}, u_{t-k}, s_{t-k+1},...,s_t, u_t]
\end{equation}
\citet{kim2021emoberta} indicated that it is difficult for the pretrained language model to distinguish the "context" (i.e., $[s_{t-k}u_{t-k}\cdots s_{t-1}u_{t-1}]$) and target turn (i.e., $s_t,u_t$). Inspired by prompt learning\cite{liu2021pre}, we construct a prompt for the $t$-th turn as follows.
\begin{equation}
    P_t = \texttt{for } u_t, s_t  \texttt{ fells <mask>}
\end{equation}
The full input of the encoder is $C_t\oplus P_t$, where $\oplus$ is the concatenation operation. In order to let the encoder realize that the prompt contains the target sentence, for the training pair of $t$-th turn  $X_t^{t}=\{C_t\oplus P_t, y_t\}$, we construct an additional training pair $X_t^{h}=\{C_t\oplus P_h, y_h\}$, where $h$ is randomly selected from $(t-k,\cdots,t)$. $X_t^{t}$ and $X_t^{h}$ shares the same context but has different prompts and labels. Training on such data helps the model to pay more attention to the target sentence and generate reasonable representations.

For a training pair $X_{t}^{k}$, we first feed $C_t\oplus P_k$ into the SimCSE model and get the last hidden states $H_t^k\in \mathbb{R}^{l\times d}$,
\begin{equation}
    H_t^{k} = \mathrm{SimCSE}(C_t\oplus P_k)
\end{equation}
where $l$ is the number of tokens in $C_t\oplus P_k$, and $d$ is the dimension of a token embedding.
Then we use the embeddings of the special token \texttt{<mask>} from $H_t^{k}$ as a representation of $y_k$-th emotion.

\subsection{Supervised Prototypical Contrastive Learning for ERC}

\paragraph{Supervised Contrastive Learning}
Supervised contrastive learning\cite{khosla2020supervised} treats all examples with the same label in the batch as positive examples. A batch of $N$ emotion representations generated via context encoder is denoted as $I = [z_1, z_2,\cdots, z_N]$. The vanilla supervised contrastive learning computes the loss  $\mathcal{L}_i^{sup}$ for $z_i$ as follows,
\begin{equation}
    \mathcal{F}(z_i,z_j) = \exp (\mathcal{G}(z_i,z_j)/\tau)
    \label{eq4}
\end{equation}

\begin{equation}
    \mathcal{N}_{sup}(i) = \sum_{z_j \in A(i)} \mathcal{F}(z_i,z_j)
    \label{eq5}
\end{equation}
\begin{equation}
    \mathcal{P}_{sup}(i) = \sum_{z_p \in P(i)} \mathcal{F}(z_i,z_p)
\end{equation}

\begin{equation}
    \mathcal{L}_i^{sup}=  -\log \frac{1}{|P(i)|}\frac{\mathcal{P}_{sup}(i)}{\mathcal{N}_{sup}(i)}
    \label{eq7}
\end{equation}
Here, $\mathcal{G}(z_i, z_j)$ is a score function that can be dot production, cosine similarity, etc. In our work, we use the cosine similarity for $\mathcal{G}(\cdot)$. $\tau\in\mathbb{R}^{+}$ is a scalar temperature parameter. $A(i) \equiv I \backslash \{z_i\}$ contains all representations in $I$ except $z_i$, and $P(i)$ is the set of positive samples that have the same label with $z_i$ in a batch.




\paragraph{Prototypical Contrastive Learning}
The object function $\mathcal{L}^{sup}$ introduces contrastive learning into supervised learning scenarios but suffers from class-imbalanced problem. Due to the limitations of batch size, samples from the majority class (e.g., \texttt{neutral} emotion) of the dataset may see insufficient negative samples within a batch. At the same time, it is hard for samples from the minority class to meet positive samples. 

To solve this issue, we design a supervised prototypical contrastive learning (SPCL) loss function, which introduces prototype vectors of each category into the $\mathcal{L}^{sup}$ loss. First, we maintain a fixed-size representation queue for each emotion category. A representation queue for $i$-th emotion with size $M$ is denoted as $Q_i=[z_1^{i},z_2^{i},\cdots, z_{M}^{i}]$. When a new representation $\hat{z}^{i}$ of $i$-th emotion is generated, we will first remove the oldest element in $Q_i$ if $|Q_i|$ equals $M$, then detach the gradient of $\hat{z}^{i}$ and push it into $Q_i$. Second, to calculate the prototype vector for $i$-th category, we randomly select $K$ samples from $Q_i$ as the support set $S_K$, then take the mean of support set as the prototype vector $\mathbf{T}_{i}$.

\begin{equation}
    S_K = \mathrm{R\textsc{andom}S\textsc{elect}}(Q_i, K)
    \label{eq8}
\end{equation}
\begin{equation}
    \mathbf{T}_{i} = \frac{1}{K}\sum_{z_{j}^{i} \in S_K,  j \in [1...K]} z_{j}^{i}
    \label{eq9}
\end{equation}

We can get at most $C_M^K$ different prototypes through sampling even if the representation queue is not updated.

After obtaining the prototype vectors, we treat each of them as an example of the corresponding category, so the sum of negative scores of $z_i$ can be calculated as follows,

\begin{equation}
    \mathcal{N}_{spcl}(i) = \mathcal{N}_{sup}(i) + \sum_{k \in \mathcal{E} \backslash y_i}\mathcal{F}(z_i, \mathbf{T}_{k})
    \label{eq12}
\end{equation}
where $y_i$ is the emotion label of $i$-th sample.
Simultaneously, the sum of positive scores of $z_i$ is also computed with the corresponding prototype vector.
\begin{equation}
    \mathcal{P}_{spcl}(i) = \mathcal{P}_{sup}(i) + \mathcal{F}(z_i,\mathbf{T}_{y_i})
    \label{eq13}
\end{equation}
Based on Eq.(\ref{eq12}) and Eq.(\ref{eq13}), the SPCL loss can be formulated as follows,
\begin{equation}
    \mathcal{L}_{i}^{spcl} = -\log \left(\frac{1}{|P(i)|+1} \cdot \frac{\mathcal{P}_{spcl}(i)}{\mathcal{N}_{spcl}(i)}\right)
    \label{eq14}
\end{equation}

The total SPCL loss of a batch is as follows,
\begin{equation}
    \mathcal{L}^{spcl} = \sum_{i=1}^{N} \mathcal{L}_{i}^{spcl}
    \label{eq15}
\end{equation}

In summary, by introducing the prototype vectors, the SPCL loss ensures that there are at least one positive pair and $|\mathcal{E}|-1$ negative pairs for each sample in a batch.

\subsection{Curriculum Learning}
\label{cl}
Existing ERC datasets are usually collected in a multi-modal fashion. When building a text-only ERC model, some utterances are not informative enough to judge the emotions. Training the model with these extreme samples will lead to performance degradation. In this paper, we try to use curriculum learning to alleviate this issue. 
\paragraph{Difficulty Measure Function} To combine with contrastive learning, we propose a difficulty measure function based on the distance between classes. Let the total size of training set $\mathcal{D}_{train}$ as $L$, the emotion representation of $i$-th data sample as $z_i$, and the label of $i$-th data sample as $y_i$. Before each training epoch, we first compute $z_i$ for all samples, then the center of $k$-th emotion is computed as follows,
\begin{equation}
    \mathbf{C}_{k} = \frac{1}{|\{z_j|\forall j,y_j=k\}|}\sum_{j=1}^{L}z_j \cdot \mathbb{I}(y_j=k)
    \label{eqck}
\end{equation}

The difficulty of $i$-th sample $\mathcal{DIF}(i)$ is calculate as follows,

\begin{equation}
    \mathcal{DIF}(i) = \frac{\mathrm{dis}(z_i, \mathbf{C}_{y_i})}{ \sum_{j=1}^{|\mathcal{E}|} \mathrm{dis}(z_i, \mathbf{C}_j)}
    \label{func:diff}
\end{equation}
$\mathrm{dis}$ function here is cosine distance. This function has the following two properties:
\begin{itemize}
    \item The closer the sample is to the category center, the lower the difficulty.
    \item For two samples with the same distance from the center within the category, the further away from the center of other categories, the lower the difficulty.
\end{itemize}


\paragraph{Curriculum Strategy} After sorting the entire training set, instead of directly splitting the training set, we design a sampling-based approach to construct a series of subsets ranging from easy to hard. Let $R$ as the number of training epochs, to train the model at $k$-th epoch, we first generate a arithmetic progression $a$ with a length of $L$, where $a_1=1-k/R$ and $a_L = k/R$. Then we initialize a Bernoulli distribution with $a$ and draw a binary random array $R_B$ from it.
We use $B$ to draw a subset $\mathcal{D}_{sub-k}$ from training set for the current epoch, where $\mathcal{D}_{sub-k} \equiv \{x_i \in \mathcal{D}_{train}| {R_B}_i = 1\}$.
Obviously, $\mathcal{D}_{sub-0}$ mainly consists of easy samples and $\mathcal{D}_{sub-R}$ mainly consists of hard samples. Compared to splitting the training set sequentially, the sampling-based approach provides a smoother difficulty variation for the model. The curriculum strategy is illustrated in  $line 2-line 9$ of Algorithm \ref{alg1}. We conduct a qualitative analysis of our curriculum learning strategy in Section \ref{cl-analysis}.
\subsection{Training and Evaluation}
\paragraph{Training} The overall procedure of our proposed approach is illustrated in Algorithm \ref{alg1}. We first generate emotion representations for all samples in training set, then use them to compute difficulty for each sample. After sorting the training set based on difficulty, we sample a subset $S_K$ and train the context encoder on $S_K$ through the SPCL loss.
\paragraph{Evaluation}  Since we computed the center of each class $\mathbf{C}$ when calculating SPCL loss, we can directly obtain the prediction through matching the centers as follows,
\begin{equation}
    p_{m}^{ic} = \frac{\mathcal{G}(z_i,\mathbf{C}_c)}{\sum_{k=1}^{|\mathcal{E}|}\mathcal{G}(z_i,\mathbf{C}_k)}
\end{equation}
where $p_{m}^{ic}$ indicates the probability that $i$-th sample belongs to category $c$, and the subscript $m$ means $p_{m}^{ic}$ is calculated through matching.

For comparison, we train an additional linear layer to predict the labels using cross-entropy loss,
\begin{equation}
    p_l^{i} = \mathrm{W}\cdot z_i + b
\end{equation}
\begin{equation}
    \mathcal{L}_{CE} = -\frac{1}{N}\sum_{i=1}^{N}\sum_{c=1}^{\mathcal{E}}y_{ic} \cdot \log p_{l}^{ic}
\end{equation}
where $\mathrm{W}\in \mathbb{R}^{dim \times |\mathcal{E}|}$ is a trainable parameter. The gradient of $z_i$ is detached so the model is only optimized via contrastive learning loss. 

In this paper,  we use $p_{m}^{i}$ to predict labels when SPCL is the loss function and use $p_{l}^{i}$ for other cases.

\begin{algorithm}[t]
\caption{Training Process with SPCL and Curriculum Learning}
\begin{algorithmic}[1]
\REQUIRE
$\mathcal{D}_{train}$:the training set with size $L$\\
$R$: the number of total epochs\\
$K$: the size of support set $S_K$\\
$M$: the context encoder \\
$\mathcal{E}$: the label set
\ENSURE the optimal model $M^*$\\
category centers $\mathbf{C}_j, j \in 1..|\mathcal{E}|$
\FOR{$k$=0 to $R$}
    \STATE $I = \{M(x_i), \forall x_i \in \mathcal{D}_{train}$\}
    \STATE compute $\mathbf{C}_j, j\in 1.. |\mathcal{E}|$ (Eq.\ref{eqck})
    \STATE compute $\mathcal{DIF}(i), i=1..L$ (Eq.\ref{func:diff})
    \STATE $\mathcal{D}_{train}$ = sort($\mathcal{D}_{train}$, $\mathcal{DIF}$)
    \STATE $st = k/R, ed=1-k/R$
    \STATE $a_1 = st, a_n=a_1 + (n-1)\cdot \frac{ed-st}{L-1}$
    \STATE $R_B \sim \mathrm{Bernoulli}(p=a)$
    \STATE $\mathcal{D}_{sub-k} \equiv \{x_i \in \mathcal{D}_{train}| {R_B}_i = 1\}$
    \STATE $Q_j$ = [],$j\in 1..|\mathcal{E}|$
    \FOR{$batch \in \mathcal{D}_{sub-k}$}
        \STATE U\textsc{pdate}($Q_j$), $j\in 1..|\mathcal{E}|$
        \STATE ${S_K}_j = \mathrm{R\textsc{andom}S\textsc{elect}}(Q_j, K)$
        \STATE compute prototype $\mathbf{T}_j$, $j\in 1..|\mathcal{E}|$ (Eq.\ref{eq9})
        \STATE compute SPCL loss (Eq.\ref{eq12}-Eq.\ref{eq15})
        \STATE optimize($M$)
    \ENDFOR
\ENDFOR
\RETURN $M^*, \mathbf{C}_j, j \in 1..|\mathcal{E}|$ 
\end{algorithmic}
\label{alg1}
\end{algorithm}

\begin{table*}[h]
\centering
\begin{tabular}{l|C{2cm}|C{2cm}|C{2cm}}
\hline
Models              & IEMOCAP        & MELD           & EmoryNLP       \\ \hline
COSMIC\cite{ghosal2020cosmic}              & 65.28          & 65.21          & 38.11          \\ \hline
DialogueCRN \cite{hu2021dialoguecrn}         & 66.46          & 63.42          & 38.91          \\ \hline
DAG-ERC \cite{shen2021directed}             & 68.03          & 63.65          & 39.02          \\ \hline
TODKAT \cite{zhu2021topic}             & 61.33          & 65.47          & 38.69          \\ \hline
Cog-BART \cite{li2021contrast}            & 66.18          & 64.81          & 39.04          \\ \hline
TUCORE-GCN\_RoBERTa\cite{lee2021graph} & -                  &  65.36          & 39.24        \\ \hline
SGED + DAG-ERC\cite{ijcai2022p562}      & 68.53          & 65.46          & 40.24          \\ \hline
EmotonFlow-Large \cite{song2022emotionflow}    & -              & 66.50          & -              \\ \hline
CoMPM \cite{lee2021compm}               & 69.46          & 66.52          & 38.93          \\ \hline
 \hline
SPCL-CL-ERC(Ours)       & $\mathbf{69.74}$ & $\mathbf{67.25}$ & $\mathbf{40.94}$ \\ \hline
\end{tabular}
\caption{Performance comparisons on three datasets.}
\label{tab:main}
\end{table*}

\section{Experimental Settings}

\subsection{Experimental Setup}
The code framework and initial weight of SimCSE come from Huggingface’s Transformers\cite{wolf2020transformers}. We use the AdamW optimizer and cosine learning rate schedule strategy. When constructing training samples, we restrict their length to less than 256. We search the hyper-parameters on the develop set. For all experiments in this paper, we keep the best checkpoint on the develop set,  then report the results on the test set using the kept checkpoint. All experiments are conducted on Nvidia V100 GPU.
\begin{table}[t]
    \centering
\begin{tabular}{c|cccc}
\hline  & MELD & IEMOCAP & EmoryNLP \\
\hline No.Dials & 1,432 & 151 & 827 \\
Train  & 1,038 & 100 & 659 \\
Dev & 114 & 20 & 89 \\
Test  & 280 & 31 & 79 \\
\hline No.Uttrs  & 13,708 & 7,333 & 9,489 \\
Train  & 9,989 & 4,810 & 7,551 \\
Dev  & 1,109 & 1,000 & 954 \\
Test  & 2,610 & 1,523 & 984 \\
\hline
No.CLS & 7 & 6 & 7 \\
\hline
\end{tabular}
    \caption{Statistics of the three datasets.}
    \label{tab:my_label}
\end{table}

\subsection{Datasets}

We conduct experiments on three widely used benchmarks: MELD\cite{poria2019meld}, EmoryNLP\cite{zahiri2018emotion} and IEMOCAP\cite{busso2008iemocap}.

\paragraph{MELD}  This dataset has more than 1400 dialogues and 13000 utterances from Friends TV series. Multiple speakers participated in the dialogues. Each utterance in a dialogue has been labeled by any of these seven emotions -- \texttt{Anger, Disgust, Sadness, Joy, Neutral, Surprise and Fear}. 

\paragraph{EmoryNLP} This dataset comprises 97 episodes, 897 scenes, and 12,606 utterances, where each utterance is annotated with one of the seven emotions borrowed from the six primary emotions in the Willcox’s feeling wheel\cite{willcox1982feeling}, i.e., \texttt{Sad, Mad, Scared, Powerful, Peaceful, Joyful}, and a default emotion of \texttt{Neutral}.

\paragraph{IEMOCAP} This dataset consists of 151 videos of recorded dialogues, with 2 speakers per session for a total of 302 videos across the dataset. Each segment is annotated for the presence of 9 emotions (\texttt{Angry, Excited, Fear, Sad, Surprised, Frustrated, Happy, Disappointed} and \texttt{Neutral}). The dataset is recorded across 5 sessions with 5 pairs of speakers.

The Statistics of these datasets are listed in Table \ref{tab:my_label}. \texttt{No.dials} stands for the number of dialogues while \texttt{No.uttrs} stands for the total number of utterances in the dataset. \texttt{No.CLS} is the number of different emotions in the dataset.
\subsection{Metrics}
From Figure \ref{fig:my_label} we can see class-imbalance in all three benchmarks, so we use weighted-F1 score as the metric for all experiments in this paper.

\section{Results and Analysis}

\subsection{Main Results}

We compare our proposed approach with state-of-the-art text-based ERC methods, and the results are presented in Table \ref{tab:main}. We can see that combining our proposed SPCL and curriculum learning strategy, we achieves state-of-the-art results on three benchmarks, which outperform previous SOTAs by 0.28\%(CoMPM on IEMOCAP), 0.73\%(CoMPM on MELD) and 0.7\%(SGED + DAG-ERC on EmoryNLP).
\subsection{Ablation Study}

To evaluate the individual effects of SPCL and CL, we conducted a series of ablation experiments, and the results are shown in Table \ref{tab:abl}. The first line in Table \ref{tab:abl} shows the performances of our proposed prompt-based context encoder trained with cross-entropy loss, which is our baseline model. We notice that curriculum learning didn't help a lot with cross-entropy loss. We believe that it is because we use the cosine distance in the difficulty measure function $\mathcal{DIF}$. However, it is unreasonable to compute cosine distance directly on representations optimized via cross-entropy loss.

 The SupCon loss performs better than cross-entropy loss on MELD and EmoryNLP datasets but slightly worse than on the IEMOCAP dataset. Combining the three results, we can see no significant performance gap between SupCon and cross-entropy losses. But the combination of curriculum learning and SupCon(SupCon+CL) shows consistent superiority since SupCon uses cosine similarity as the score function. For the representations generated by SupCon, the cosine distance between representations of the same category will be closer, and the distance between representations of different categories will be distant. Therefore, the difficulty measure function of CL can be more faithful, resulting in better performance.
 
 The SPCL loss outperforms SupCon and cross-entropy losses on all three datasets. Meanwhile, it also has consistent performance improvements in combination with curriculum learning.

To summary, both our proposed SPCL and curriculum learning strategy contribute significantly to the results.
\begin{table}[t]
\begin{tabular}{l|C{1.6cm}|C{1.6cm}|C{1.6cm}}
\hline
& IEMOCAP & MELD  & EmoryNLP \\ \hline

CE                   & 68.35        & 65.33      &   38.72       \\ 
\quad + CL                   & 67.40        & 65.63      &   39.00       \\ \hline
SupCon                     & 68.13        &   65.67    &  39.20        \\ 
\quad + CL                     & 68.64        &   66.15    &  39.49        \\ \hline
SPCL                      &    69.03     & 66.56     &     40.14     \\ 
\quad + CL                     & 69.74   & 67.25 & 40.94    \\ \hline
\end{tabular}
\caption{Results of ablation study. Here, CE means Cross-entropy loss, SupCon is the vanilla supervised contrastive learning loss and SPCL is our proposed supervised prototypical contrastive learning loss. CL is our proposed curriculum strategy.}
\label{tab:abl}
\end{table}

\begin{figure}[h]
    \centering
    \includegraphics[width=0.48\textwidth]{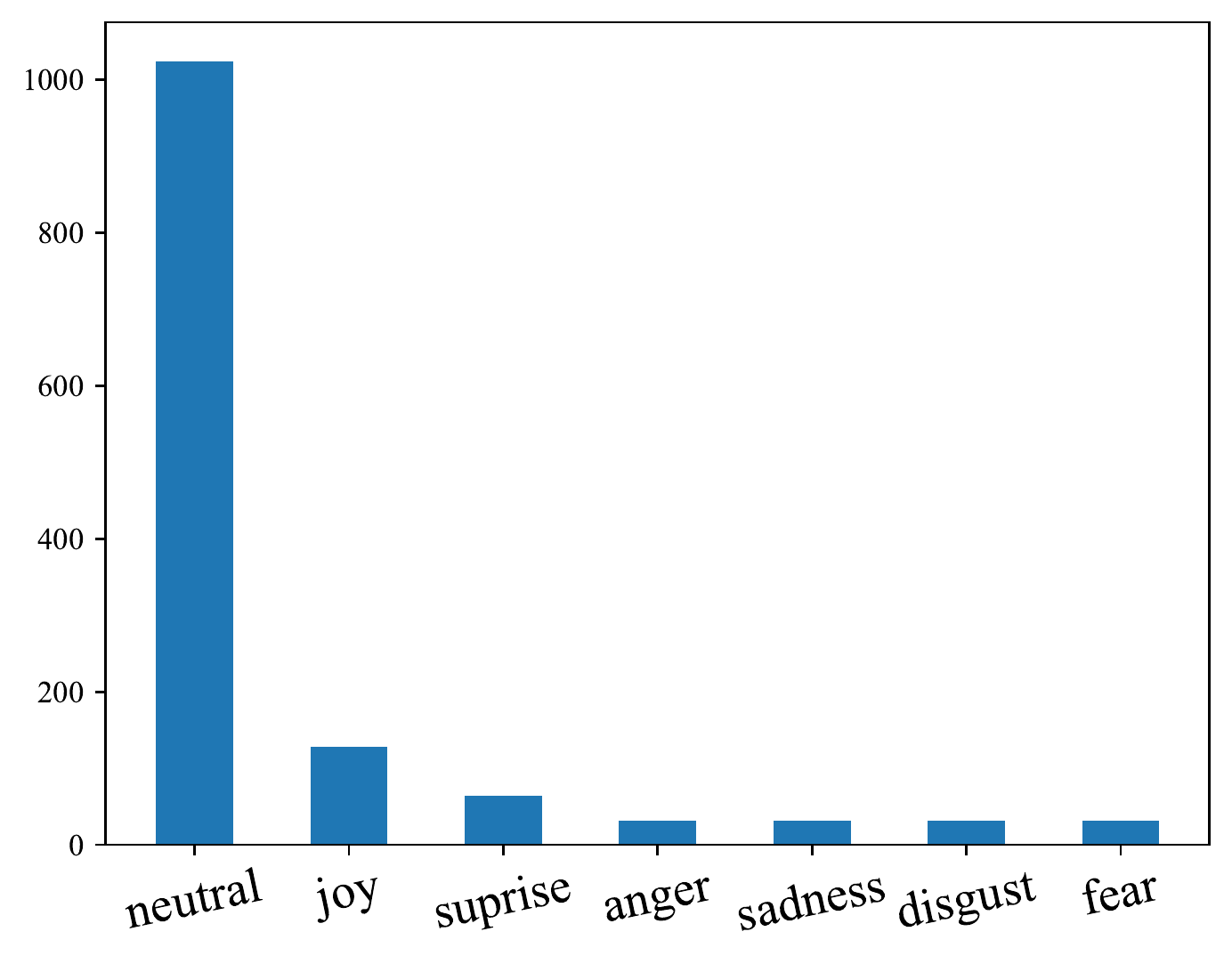}
    \caption{Emotion distribution of the extreme class-imbalanced training set. We construct it from MELD training set.}
    \label{fig:imb}
\end{figure}

\subsection{Using SPCL on Class Imbalanced Data}
\label{sec:imb}
To demonstrate the superiority of SPCL on imbalanced data, we construct an imbalanced subset from MELD training set, as illustrated in Figure \ref{fig:imb}. We sample 1024, 128, 64, 32, 32, 32, and 32 samples of \texttt{neutral, joy, surprise, anger, sadness, disgust}, and \texttt{fear}, respectively. We train the model using these two loss functions on the imbalanced training set. Since a small batch size will aggravate the impact of class-imbalance on contrastive learning, we conducted experiments with batch size of 4, 8, 16, and 32, respectively. The results on MELD test set are shown in Table \ref{tab:imb}. We notice that using SPCL outperforms using SupCon in all four sets of experiments. As the batch size decreases from 32 to 4, the weighted-F1 score of SupCon loss drops 6.95\% while SPCL drops 4.1\%. We can conclude that in the class-imbalance scenarios: 1)  both SupCon and SPCL need a larger batch size to reach satisfied performances; 2) introducing the prototypical network into contrastive learning can alleviate the impact of class-imbalance.

\begin{table}[h]
\begin{tabular}{l|C{1.1cm}|C{1.1cm}|C{1.1cm}|C{1cm}} \hline
       & 4     & 8     & 16    & 32    \\ \hline
SupCon & 53.14 & 57.36 & 58.50 & 60.09 \\ \hline
SPCL   & 57.27 & 58.85 & 59.47 & 61.38 \\ \hline
\end{tabular}
\caption{Results of different loss functions and different batch sizes trained on the imbalanced training set.}
\label{tab:imb}
\end{table}

\subsection{Using SPCL with Small Batch Size }

Contrastive learning approaches usually need a large batch size to ensure more positive/negative pairs within a batch, which leads high computational cost. In the Section \ref{sec:imb}, we find that both SupCon and SPCL relay on large batch sizes. We conjecture that SPCL's dependence on batch size may be because we sample too few data samples for some categories(i.e., 32 for \texttt{anger, sadness,disgust, and fear}) to compute reasonable prototypes at the beginning of the training. In order to further investigate the effect of batch sizes on SPCL, we apply SPCL to a more general scenario. From Figure \ref{fig:my_label} we can see the IEMOCAP dataset is not extreme class-imbalanced. Even the smallest category (\texttt{happy}) still has hundreds of samples. We conduct experiments on the IEMOCAP dataset, and the results are illustrated in Table \ref{tab:batchsize}. Experimental results show that given enough samples for each category, the SupCon loss still needs a large batch size, but the SPCL doesn't. With the batch size decreasing from 32 to 4, the performance of SupCon drops 5.63\%, while SPCL only drops 0.82\%.This demonstrates that the SPCL loss is less sensitive to the training batch size.

\begin{table}[h]
\begin{tabular}{l|C{1.1cm}|C{1.1cm}|C{1.1cm}|C{1cm}} \hline
       & 4     & 8     & 16    & 32    \\ \hline
SupCon & 62.50 & 65.01 & 67.04 & 68.13 \\ \hline
SPCL   & 68.21 & 68.41 & 68.48 & 69.03 \\ \hline
\end{tabular}
\caption{Results of SupCon and SPCL with different batch sizes on IEMOCAP dataset.}
\label{tab:batchsize}
\end{table}

\subsection{Qualitative Analysis of Curriculum Learning}
\begin{figure}[h]
    \centering
    \includegraphics[width=0.48\textwidth]{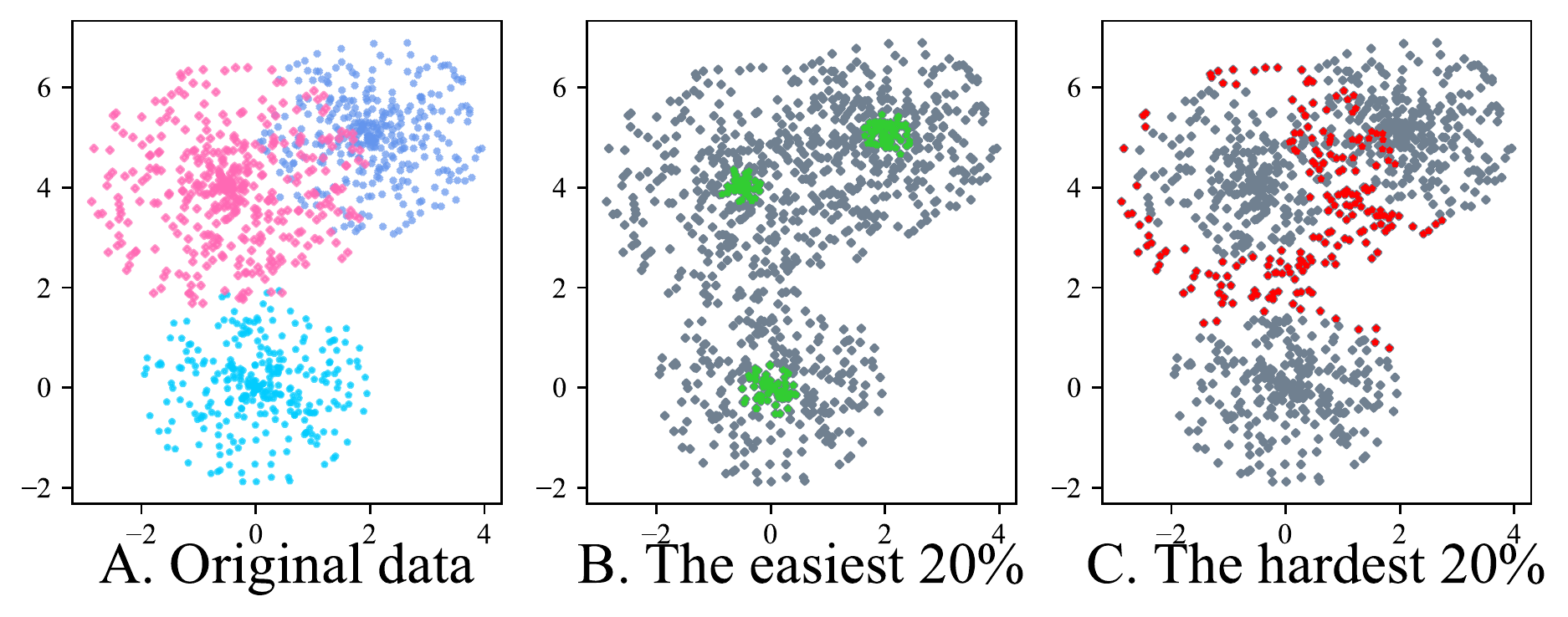}
    \caption{Visualizations of how the difficulty measure function $\mathcal{DIF}$ in Eq.(\ref{func:diff}) ranks the data.}
    \label{fig:diffcult}
\end{figure}

\label{cl-analysis}
To conduct a qualitative analysis of our proposed curriculum learning strategy, we generate a toy dataset that contains three classes and visualize it in Figure \ref{fig:diffcult}(A).

As illustrated in Figure \ref{fig:diffcult}(B-C), the difficulty measure function $\mathcal{DIF}$ ranks the samples in a reasonable way. The easiest 20\% samples(in green) are distributed in the center of their respective categories, while the hardest 20\% samples(in red) are mainly on the boundaries between classes. 

\begin{figure}[t]
    \centering
    \includegraphics[width=0.48\textwidth]{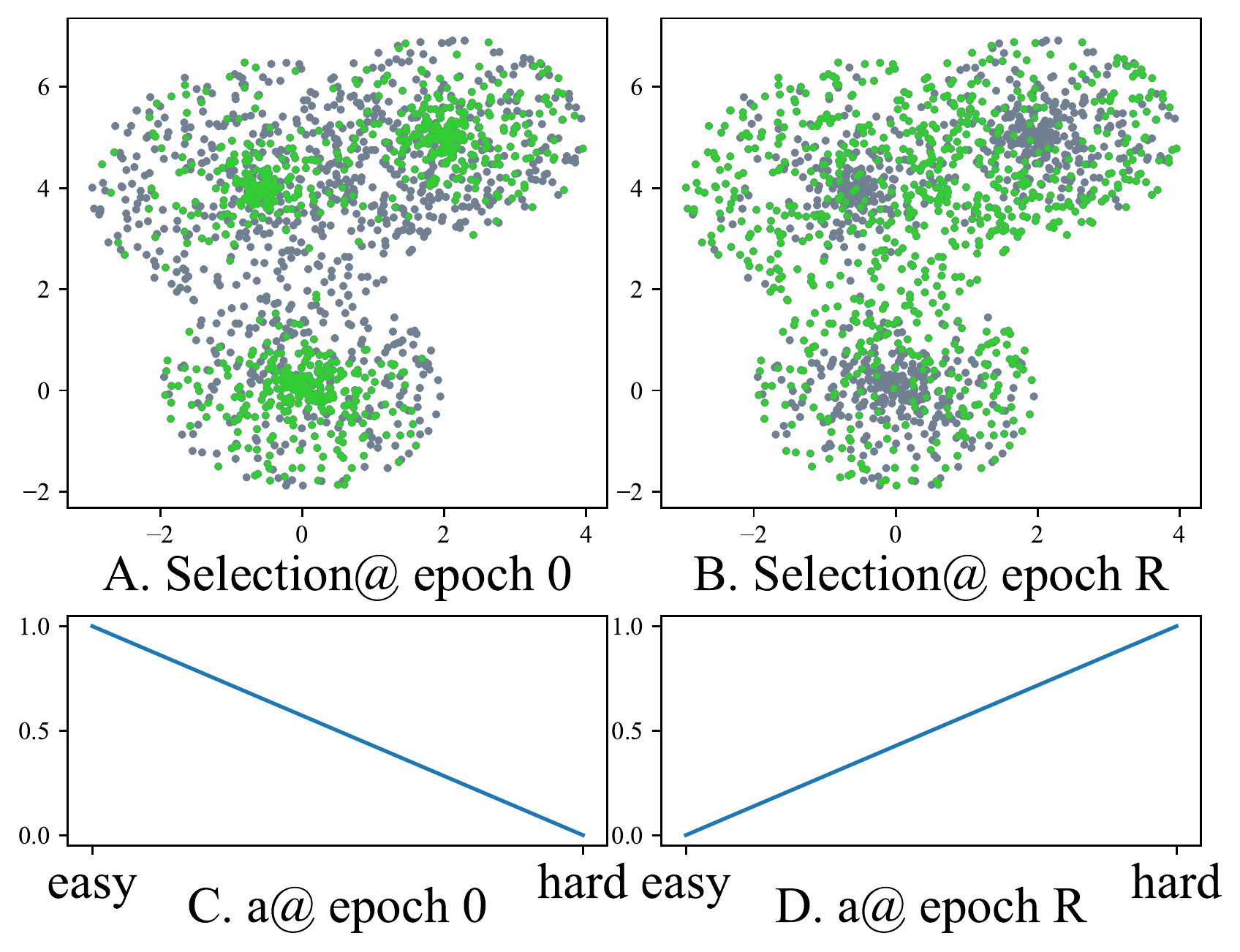}
    \caption{The sampling-based curriculum strategy.}
    \label{fig:selection}
\end{figure}

In practice, we found that directly sorting the data with $\mathcal{DIF}$ cannot obtain satisfactory results. The model will overfit on simple samples in the early stage of training and produce large losses in the later stage, so we design the sampling-based curriculum learning strategy described in Section \ref{cl} to provide a smoother difficulty variation for the model. We control the difficulty of training subsets based on the sampling probability of samples with different difficulties.  As illustrated in Figure \ref{fig:selection}, the arithmetic progression $a$ we used to sample downs from 1 to 0 at the first epoch, the sampling results are shown in Figure \ref{fig:selection}(A), we can find that the most selected samples(in green) are around the centers while a few samples are away from the centers. When running to the last epoch, $a$ grows from 0 to 1, so hard samples are in the majority of the subset, as shown in Figure \ref{fig:selection}(B).

\section{Conclusion}

In this paper, we propose a novel loss function called \textbf{S}upervised \textbf{P}rototypical \textbf{C}ontrastive \textbf{L}earning (\textbf{SPCL}) loss for the emotion recognition in conversation task. Combining with Prototypical Network, the SPCL loss outperforms the traditional supervised contrastive learning loss. It also works well on class-imbalanced data and is less sensitive to the training batch size, which reduces the requirement of computing resource. To further exploit the power of contrastive learning on ERC tasks, we design a distance-based difficulty measure function and introduce curriculum learning to alleviate
the impact of extreme samples. We conduct experiments on three widely used benchmarks: IEMOCAP, MELD, and EmoryNLP. Results show that our approach achieves state-of-the-art performance on all three datasets.

\section*{Limitations}

This work has three limitations: 1) We introduce too many hyperparameters, which requires additional computing resources to search. 2) Our proposed difficulty measure function can not be combined with most existing ERC methods since it requires the emotion representations produced by the ERC model to be distance-aware. 3) We used multiple random sampling, resulting in unstable performance. The results in this paper are averaged with multiple seeds. In practice, we found that the results generated by different seeds may have significant variance.
\bibliography{anthology,custom}
\bibliographystyle{acl_natbib}




\end{document}